\documentclass[letterpaper, 10 pt, conference]{ieeeconf}  

\IEEEoverridecommandlockouts                              

\usepackage{cite}

\usepackage{booktabs}
\usepackage{multirow}
\usepackage{amsmath}
\usepackage{pifont}
\usepackage{makecell}
\usepackage{float}
\usepackage{graphicx}
\usepackage{amssymb}
\usepackage{threeparttable}
\usepackage{url}

\DeclareMathOperator*{\argmin}{arg\,min}

\title{\LARGE \bf
RVN-Bench: A Benchmark for Reactive Visual Navigation
}

\author{Jaewon Lee$^{1}$, Jaeseok Heo$^{2}$, Gunmin Lee$^{2}$, Howoong Jun$^{3}$, Jeongwoo Oh$^{4}$ and Songhwai Oh$^{1}$
\thanks{$^{1}$J. Lee, J. Heo, G. Lee and S. Oh are with the Department of Electrical and Computer Engineering, Seoul National University (SNU) and Automation and Systems Research Institute (ASRI) and Sequor Robotics Inc., Seoul, Korea (Republic of).
        {\tt\small jaewon.lee@rllab.snu.ac.kr, jaeseok.heo@rllab.snu.ac.kr, songhwai@snu.ac.kr}}%
\thanks{$^{2}$Gunmin Lee is with the Department of Electrical and Computer Engineering, Seoul National University (SNU) and Automation and Systems Research Institute (ASRI), Seoul, Korea (Republic of).
        {\tt\small gunmin.lee@rllab.snu.ac.kr}}%
\thanks{$^{3}$Howoong Jun is with Interdisciplinary Program in Artificial Intelligence, Seoul National University (SNU) and Automation and Systems Research Institute (ASRI) and Sequor Robotics Inc., Seoul, Korea (Republic of).
        {\tt\small howoong.jun@rllab.snu.ac.kr}}%
\thanks{$^{3}$Jeongwoo Oh is with Sequor Robotics Inc., Seoul, Korea (Republic of).
        {\tt\small jeongwoo.oh@sequorrobotics.com}}%
\thanks{Corresponding author: Songhwai Oh}
}

\begin{document}

\maketitle
\thispagestyle{empty}
\pagestyle{empty}

\begin{figure*}[t]
    \centering
    \includegraphics[width=\linewidth]{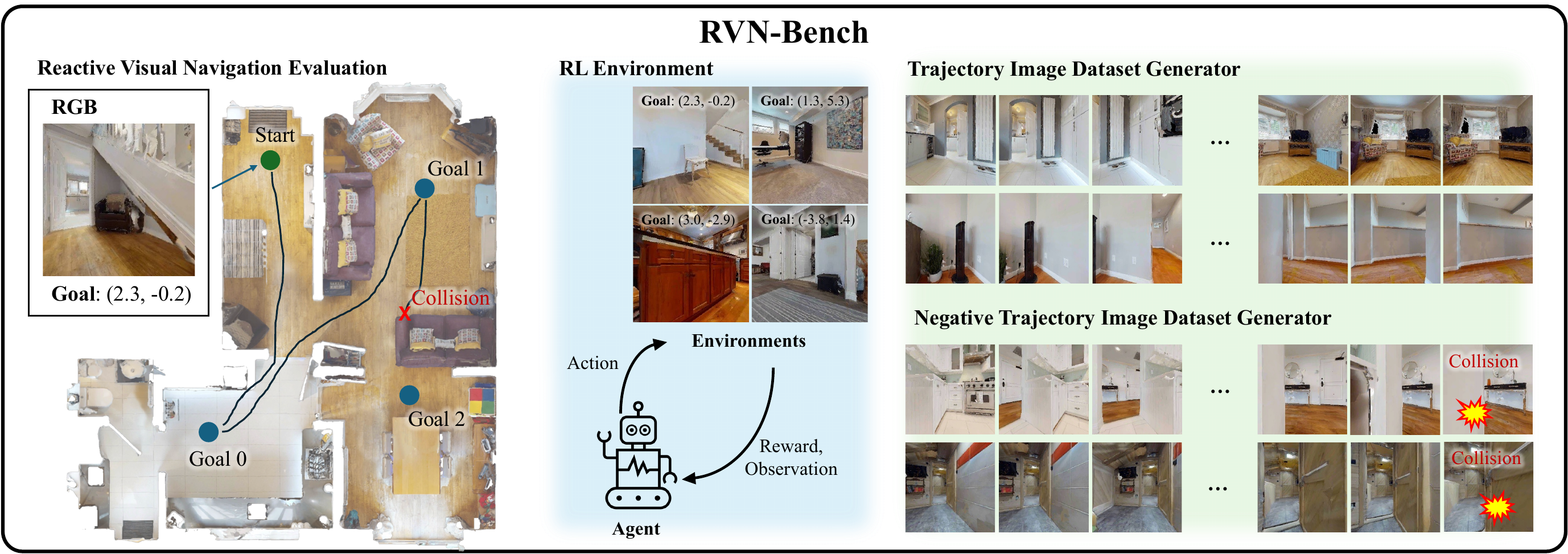}
    \caption{\textbf{Overview of the RVN-Bench.} The benchmark is designed for indoor mobile robots and focuses on collision-aware visual navigation, where an agent must reach sequential goal positions using only visual observations while avoiding collisions. Built upon Habitat 2.0 and utilizing HM3D scenes, it provides high-quality visual observations, an RL environment for training and evaluation, and pipelines for collecting trajectory image datasets, including negative (collision-inducing) trajectories.}
    \label{fig:overview}
\end{figure*}

\begin{abstract}

Safe visual navigation is critical for indoor mobile robots operating in cluttered environments.
Existing benchmarks, however, often neglect collisions or are designed for outdoor scenarios, making them unsuitable for indoor visual navigation.
To address this limitation, we introduce the reactive visual navigation benchmark (RVN‑Bench), a collision-aware benchmark for indoor mobile robots.
In RVN‑Bench, an agent must reach sequential goal positions in previously unseen environments using only visual observations and no prior map, while avoiding collisions.
Built on the Habitat 2.0 simulator and leveraging high‑fidelity HM3D scenes, RVN‑Bench provides large‑scale, diverse indoor environments, defines a collision‑aware navigation task and evaluation metrics, and offers tools for standardized training and benchmarking.
RVN‑Bench supports both online and offline learning by offering an environment for online reinforcement learning, a trajectory image dataset generator, and tools for producing negative trajectory image datasets that capture collision events.
Evaluations demonstrate that policies trained on RVN-Bench generalize effectively across unseen simulated environments. Furthermore, initial physical experiments using a Jackal UGV indicate promising sim-to-real transfer.
Code and additional materials are available at: \url{https://sequor-robotics-research.github.io/projects/RVN-Bench/}.

\end{abstract}

\section{Introduction}


Reactive visual navigation (RVN) is the problem of reaching specified goals while avoiding collisions with obstacles in previously unseen environments using only visual observations, without relying on a prior map or task-specific knowledge.
RVN is critical for autonomous mobile robots, which must maintain safety under unexpected environmental changes. 
Despite recent advances in visual navigation foundation models \cite{shah2023gnm, shah2023vint, sridhar2024nomad, liu2025citywalker} that have shown promising performance on reactive visual navigation tasks, the problem of ensuring safety, particularly in obstacle-rich indoor settings where collisions are likely, remains unsolved.
Moreover, these approaches typically require both a massive amount of training data and reliable mechanisms for safe evaluation.
Collecting such data directly in the real world is costly and time-consuming, while real-world evaluation is often unsafe, as unverified algorithms risk damaging property or degrading hardware.
These challenges motivate the development of simulation frameworks that support scalable training and collision-aware evaluation of RVN.

To address these challenges, a wide range of simulation environments have been utilized to support the design and evaluation of navigation strategies \cite{dosovitskiy2017carla, cai2020summit, savva2019habitat, szot2021habitat, habitat2020sim2real, vuong2024habicrowd, khanna2024goat, anderson2018vision}.  These simulators provide an environment for the robot agent to learn in and benchmark the performance of the goal reaching ability of the robot agent.
However, most of these benchmarks either are designed specifically for autonomous driving and outdoor navigation, making them unsuitable for indoor mobile robots, or focus solely on whether the agent reaches the designated goal while disregarding any collisions that may occur during navigation.
For instance, CARLA \cite{dosovitskiy2017carla}, SUMMIT \cite{cai2020summit}, and MetaUrban \cite{wu2025metaurban} are unsuitable for indoor mobile robots, as CARLA and SUMMIT are designed for autonomous cars, whereas MetaUrban is designed for outdoor micromobility navigation. The Habitat Challenge \cite{habitat2020sim2real} and GOAT-Bench \cite{khanna2024goat} completely ignore collisions, while HabiCrowd \cite{vuong2024habicrowd} considers only pedestrian collisions and neglects static obstacles.
Benchmarks that ignore collisions do not fully capture the requirements of real-world navigation, yielding policies that may appear effective in simulation but could prove unsafe or impractical in real indoor environments cluttered with obstacles.
The availability of a collision‑aware simulation environment for both model development and comparative evaluation would enable researchers to systematically assess and refine navigation policies under realistic collision scenarios.

To this end, we introduce the reactive visual navigation benchmark (RVN-Bench), a benchmark for indoor mobile robots that prioritizes evaluating a robot's ability to avoid collisions while navigating toward a goal in previously unseen environments. Unlike prior benchmarks that either disregard collisions or focus on autonomous driving scenarios, RVN-Bench is designed specifically for indoor mobile robots and emphasizes safe navigation as a primary evaluation criterion. In RVN-Bench, the agent must rely solely on continuous image observations to navigate toward each successive goal point, while avoiding collisions with walls or other obstacles. Built upon the Habitat simulator \cite{savva2019habitat} for efficient execution, RVN-Bench incorporates scenes from the HM3D dataset \cite{hm3d}, which is derived from real indoor environments, to provide high-quality image observations across diverse indoor settings.
RVN-Bench is primarily intended for indoor ground mobile robots with a forward-facing RGB camera, which matches common indoor platforms.

RVN-Bench provides three main functionalities: (1) a standardized environment for benchmarking and evaluation with collision-aware metrics, (2) an interactive environment for training agents via reinforcement learning (RL), and (3) a trajectory image dataset generator that supports offline learning methods and can produce negative trajectory image datasets, in which trajectories end in collisions that are expensive to collect in real-world settings. Figure~\ref{fig:overview} presents an overview of the benchmark.

In addition, we implement baseline models using methods such as RL \cite{wijmans2020dd, schulman2017proximal}, safe reinforcement learning (Safe-RL) \cite{ray2019benchmarking}, and imitation learning (IL), including approaches aligned with GNMs \cite{shah2023vint, sridhar2024nomad}, to establish the benchmark. Through experiments, we demonstrate that the training environments provided by RVN-Bench are effective in learning generalizable visual navigation policies.

Our contributions are summarized as follows:
\begin{itemize}
\item We introduce RVN‑Bench, a novel evaluation framework for measuring reactive visual navigation ability of an agent.
\item RVN-Bench provides an environment for RL–based training.
\item RVN-Bench includes an offline data generation pipeline that produces trajectory image datasets.
\item We evaluate baseline models across RL, Safe-RL, and IL, trained with visual observations.
\end{itemize}

\begin{table*}[t]
    \caption{Comparison of Visual Navigation Benchmarks.}
    \label{table:compare_visual_bench}
    \centering
    \begin{threeparttable}
    \begin{tabular}{ccccccc}
    \toprule
    Benchmark & Goal Type & Domain & \makecell{Realistic Visual\\Rendering} & \makecell{Detect\\Collision} & \makecell{Lifelong\\Evaluation} & \makecell{Dynamic\\Obstacles} \\
    \midrule
    CARLA \cite{dosovitskiy2017carla}   & High-level command       & Autonomous driving   & \ding{51} & \ding{51}  &           & \ding{51} \\
    \midrule
    MetaUrban \cite{wu2025metaurban}    & Point                    & Outdoor              &           & \ding{51}  &           & \ding{51} \\
    \midrule
    Habitat Challenge \cite{savva2019habitat, habitatchallenge2023} & Point, object, image & Indoor & \ding{51} &            &           &           \\
    \midrule
    HM3D-OVON \cite{yokoyama2024hm3d}   & Language                 & Indoor               & \ding{51} &            &           &           \\
    \midrule
    GOAT-Bench \cite{khanna2024goat}    & Object, image, language  & Indoor               & \ding{51} &            & \ding{51} &           \\
    \midrule
    HabiCrowd \cite{vuong2024habicrowd} & Point, object            & Indoor               & \ding{51} & \ding{115} &           & \ding{51} \\
    \midrule
    \textbf{RVN-Bench (Ours)}           & Point                    & Indoor               & \ding{51} & \ding{51}  & \ding{51} &           \\
    \bottomrule
    \end{tabular}
    \begin{tablenotes}[flushleft]
        \footnotesize
        \item \ding{51}: supported; \ding{115}: collision detection is partial (pedestrian-only; static obstacles not considered); blank: not supported.
    \end{tablenotes}
    \end{threeparttable}
\end{table*}

\section{Related Work}

\subsection{Visual Navigation Benchmarks}
Visual navigation benchmarks measure agent’s ability to reach the goal given the visual observation. However, most of those benchmarks focus on measuring the agent's ability to reach the goal without explicitly penalizing collisions \cite{habitat2020sim2real, khanna2024goat, vuong2024habicrowd, yokoyama2024hm3d}. Table \ref{table:compare_visual_bench} presents a comparison between different visual navigation benchmarks. The PointNav task in the Habitat Challenge \cite{habitat2020sim2real} evaluates an agent's ability to navigate towards a goal defined by a relative position from the agent's current location.
The HabiCrowd \cite{vuong2024habicrowd} extends this to environments with pedestrians. GOAT-Bench \cite{khanna2024goat} addresses the lifelong navigation problem, where agents must follow a sequence of multi-modal instructions. However, both Habitat Challenge and GOAT-Bench disregard collisions entirely, while HabiCrowd considers only collisions with pedestrians, ignoring static obstacles such as walls and furniture. In contrast, RVN-Bench explicitly targets collision-aware navigation, requiring the agent to reach the goal while avoiding all obstacles in the environment.

\subsection{Collision-Aware Navigation Benchmarks}
There are benchmarks focusing on measuring agent’s ability to avoid collisions with obstacles\cite{ray2019benchmarking, xiao2022autonomous, dosovitskiy2017carla, cai2020summit, wu2025metaurban}. A unified safe reinforcement learning benchmark (Safety-Gym) \cite{ray2019benchmarking} provides an environment where agents must complete tasks while avoiding both static and dynamic obstacles. BARN \cite{xiao2022autonomous} assesses an agent’s ability to navigate through dense obstacle fields in both simulation and real-world settings. However, both Safety-Gym and BARN rely on LiDAR observations, rather than visual inputs. In contrast, RVN-Bench focuses on the field of visual navigation where agents have to navigate only given visual observations without LiDAR observations.

CARLA \cite{dosovitskiy2017carla} and SUMMIT \cite{cai2020summit}, designed for autonomous driving in photorealistic outdoor environments, are unsuitable for indoor mobile robot navigation. MetaUrban \cite{wu2025metaurban} evaluates micromobility navigation in synthetic outdoor environments, focusing on avoiding both static and dynamic obstacles. While it supports visual observations, the environment is procedurally generated using rule-based methods and does not include indoor scenes. RVN-Bench, on the other hand, leverages the high-fidelity, real-world indoor 3D environments provided by HM3D, offering more realistic visual inputs for indoor navigation tasks.

\subsection{Visual Navigation Methods}
Visual navigation has been extensively studied across various domains, including indoor and outdoor mobile robotics \cite{shah2023gnm, shah2023vint, sridhar2024nomad}, autonomous driving \cite{thorpe1988vision, chen2020learning}, and aerial robotics \cite{lin2021autonomous}.
Within this line of work, GNMs \cite{shah2023gnm, shah2023vint, sridhar2024nomad} have emerged as general-purpose, goal-conditioned visual navigation policies trained on diverse datasets collected from multiple robot platforms. These models are capable of controlling various robots in a zero-shot manner and can be efficiently fine-tuned for new robots or downstream tasks.
Among them, ViNT \cite{shah2023vint} employs an EfficientNet encoder to generate input tokens, which are then processed by a Transformer to predict future actions.
NoMaD \cite{sridhar2024nomad} similarly uses an EfficientNet encoder to produce tokens, which are input into a Transformer to generate context. Finally, a diffusion policy is applied to sample future actions from this context.

Both ViNT and NoMaD are trained using imitation learning techniques, leveraging large-scale datasets of expert demonstrations to learn navigation policies. In this study, we adopt ViNT and NoMaD as baseline models to evaluate and compare their performance within the RVN-Bench framework. This choice enables us to assess the effectiveness of these state-of-the-art IL models in navigating previously unseen environments while avoiding collisions, and to compare their performance against RL-based methods.

\section{Reactive Visual Navigation Benchmark}

\begin{figure*}[t]
    \centering
    \includegraphics[width=\linewidth]{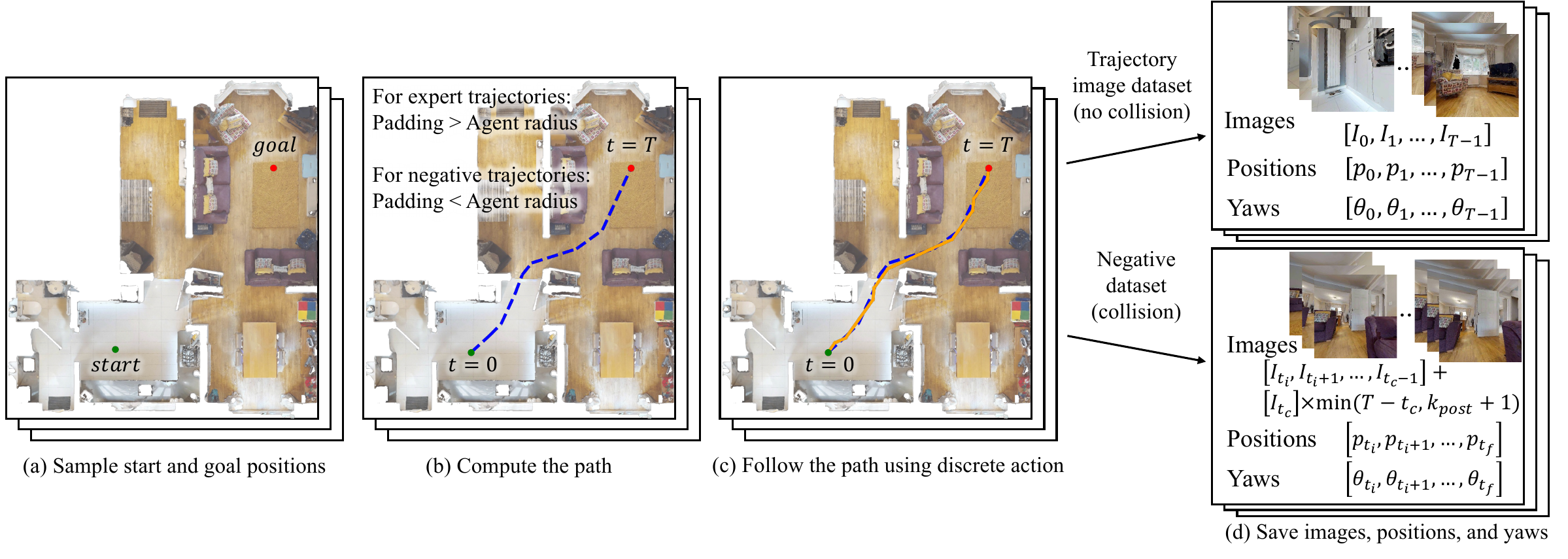}
    \caption{\textbf{Trajectory image dataset generation process.} We generate trajectory image datasets through the following steps. \textbf{(a)} Start and goal positions are randomly sampled in the scene. \textbf{(b)} The shortest path between these positions is computed using the ground truth occupancy map padded with a margin larger than the agent’s radius. \textbf{(c)} The agent follows this path using discrete actions. \textbf{(d)} If the agent reaches the goal position without collision, its positions, yaws, and image observations at each time step are recorded.
    \textbf{For negative trajectory image dataset generation}, the occupancy map is padded with a margin smaller than the agent radius to create path that is unsafe for the agent in step \textbf{(b)}. If the agent collides with an obstacle at $t=t_c$, the history of its positions, yaws, and image observations from $t_i$ to $t_f$ is recorded, where $t_i=\max(0, t_c-k_\text{pre})$, $t_f=\min(T, t_c+k_\text{post})$, and $I_t$ is set to $I_{t_c} \text{for } t>t_c$.
    }
    \label{fig:dataset_generation}
\end{figure*}

RVN-Bench is a simulation-based benchmark that explicitly targets collision-aware visual navigation for indoor mobile robots.
The benchmark defines a task in which an agent must reach sequential goal positions using only visual observations while avoiding collisions with obstacles.
The benchmark is built on the Habitat 2.0 simulator \cite{szot2021habitat} and leverages the HM3D dataset \cite{hm3d}, providing high-quality RGB observations across realistic and diverse indoor scenes.
On top of this foundation, RVN-Bench defines a standardized reactive visual navigation task, specifies evaluation metrics, and integrates tools for both training and dataset generation.
Concretely, it contributes three key functionalities: (1) a benchmarking environment with collision-aware evaluation metrics, (2) an interactive environment for training reinforcement learning (RL) agents, and (3) a trajectory image dataset generator for offline and imitation learning, supporting the creation of negative datasets where trajectories end in collisions, which are expensive to collect in the real world.
The benchmark utilizes 800 scenes for training, 50 for validation, and 50 for testing, ensuring a clear division between development and evaluation.
RVN-Bench is designed for indoor ground mobile robots with RGB sensing. It can be adapted to different robot scales by configuring the robot footprint parameters (radius/height) and motion discretization (forward step size/turn angle) for collision checking and control. The benchmark also supports configurable camera extrinsics and intrinsics (resolution, field of view, and focal length).

In the following subsections, we first define the navigation task and the agent’s inputs and outputs, then describe the RL environments, and finally present the trajectory dataset generation process, including the creation of negative trajectory image dataset.

\subsection{Task Definition} \label{taskdef}
In the RVN-Bench, the objective of an agent is to navigate to a given goal location using only the visual observations, without colliding with any obstacles.
The agent receives past and current RGB observations $\mathbf{I}_t := I_{t-C:t}$ where $I_t \in \mathbb{R}^{H\times W \times 3}, \forall t=\{0,...,T-1\}$, C represents the number of past observations, and $W$, $H$ represent the width and height of the observed image, respectively. $T$ is the total number of time steps in an episode. Along with image observations, the agent is given the goal position specified relative to its current pose, $P_t = (x_g, y_g)$.
The agent operates under nonholonomic constraints and selects an action $a_t \in \mathcal{A}=\{\texttt{MOVE\_FORWARD} (d_\text{step}), \texttt{TURN\_LEFT} (\theta_\text{step}),\\ \texttt{TURN\_RIGHT} (\theta_\text{step}), \texttt{STOP}\}$. 
The agent is modeled as a cylindrical robot with a radius of $r_\text{robot}$ and a height of $h_\text{robot}$. 
The $I_t$ is captured by a forward-facing camera mounted on the robot at a height of $h_\text{camera}$.

At the start of a new episode, the agent is spawned at a random pose in the scene, and a new goal position, which is randomly selected under certain rules, is provided. The goal is selected such that the geodesic distance from the agent’s position to the goal position lies within a specified range $[d_\text{min}, d_\text{max}]$. When the agent outputs the \texttt{STOP} action within $2 \times r_\text{robot}$ of the goal position, a new goal is given without resetting the scene or the agent’s location.
The episode is reset under three conditions below.
\begin{enumerate}
    \item Failure due to collision: the agent collides with an obstacle. 
    \item Failure due to timeout: The agent fails to reach the next goal position within $T_\text{max}$ steps.
    \item Success: The agent successfully reaches $N_\text{goal}$ goal positions. 
\end{enumerate}
Here, $T_\text{max}$ denotes the maximum episode length, and $N_\text{goal}$ the number of sequential goals defined per episode.

Collision detection relies on a precomputed navigation mesh (NavMesh) generated from scene geometry \cite{savva2019habitat}.
Upon a $\texttt{MOVE\_FORWARD}$ action, the simulator projects the target position onto the NavMesh, constraining the endpoint to the navigable surface while respecting the agent radius $r_\text{robot}$.
A collision is registered if the resulting displacement is less than the commanded step size $d_\text{step}$, indicating that the agent’s path was blocked by a NavMesh boundary.
This interaction is modeled as a hard kinematic constraint; the agent cannot penetrate boundaries, and no physics-based contact dynamics are simulated.
By incorporating an agent-radius margin during generation, the NavMesh effectively represents the configuration space of static obstacles, including complex geometries like furniture.

\subsection{Reinforcement Learning Environment}

We provide a reinforcement learning environment, facilitating the training of RL agents using online RL methods. During training, the agent is spawned in a randomly selected scene from the training set, and if the episode is reset a specified number of times, a new scene is selected. 
The agent receives a reward signal commonly used in PointGoal navigation tasks \cite{wijmans2020dd, yokoyama2024hm3d}, with an additional condition for collisions. Specifically, the agent receives a reward of $r_{\text{goal\_reached}}$ (set to 1.0) upon reaching the goal position, a terminal reward of $r_{\text{collision}}$ (set to -0.1) if a collision occurs, and a step-wise reward of $-\Delta_{\text{dtg}} - 0.01$ otherwise, where $-\Delta_{\text{dtg}}$ represents the change in the agent’s geodesic distance to the goal position. This reward structure encourages the agent to minimize the path length to the goal while penalizing collisions. For Safe-RL training, a cost of 1.0 is additionally assigned when the agent collides with an obstacle.

\[
r_t = \begin{cases}
  r_{\text{goal\_reached}} & \text{if a goal is reached} \\
  r_{\text{collision}} & \text{if a collision occurs} \\
  -\Delta_{\text{dtg}} - 0.01 & \text{otherwise}
\end{cases}
\]

\subsection{Trajectory Dataset Generation} \label{traj_data_generation}

We provide functionality to create trajectory image datasets for training offline visual navigation methods. 
The dataset follows the format of the GNM dataset \cite{shah2023gnm} and is generated as follows.
First, the start position and the goal position are randomly sampled in the RVN-Bench environment. 
The shortest path between the start and the goal position is then computed using the A* algorithm \cite{hart1968formal} based on the ground truth occupancy map of the scene, which is padded with a margin exceeding the radius of the agent. This margin ensures that the resulting path remains safe when the agent executes discrete actions.
The path is discarded if the length of the path does not lie within a specific range $[d_\text{min}, d_\text{max}]$. 
After the path is computed, the agent follows the path using discrete actions ($\mathcal{A}$) described in Section \ref{taskdef}. 
If the agent successfully reaches the goal without colliding with any obstacles, we record the history of the agent's positions, yaws, and image observations at each step as $$\mathcal{D}_\text{expert}=\{(p_t, \theta_t, I_t)\}_{t=0}^{T-1},$$ where $p_t, \theta_t$ are the position and yaw of the agent at time $t$ and $T$ is the number of time steps.
This data collection process is illustrated in Figure \ref{fig:dataset_generation}.

We also provide functionality to create a negative trajectory image dataset. 
We define a negative trajectory image dataset as a collection of positions, yaws and image observations from trajectories, where the trajectory ends in a collision. 
To collect this dataset, the shortest path between the agent’s initial position and the goal position is computed using the ground truth occupancy map, which is padded with a margin smaller than the agent’s radius. 
When the agent collides with an obstacle at $t=t_c$ while following its path, we record the partial history of the agent’s positions, yaws, and image observations as $$\mathcal{D}_\text{negative}=\{(p_t, \theta_t, I_t)\}_{t=t_i}^{t_f},$$ where $t_i=\max(0, t_c-k_\text{pre})$, $t_f=\min(T, t_c+k_\text{post})$, and $I_t = I_{t_c} \text{for } t>t_c$.
Here, $k_\text{pre}$ and $k_\text{post}$ denote the numbers of time steps recorded before and after the collision, respectively.
The dataset thus contains $k_\text{pre}$ pre-collision steps and $k_\text{post}+1$ post-collision steps.
In our experiments, we set $k_\text{pre}=8$ and $k_\text{post}=6$.
Note that the image observation remains identical to the collision-frame image for all subsequent steps $t>t_c$.
This data collection process is illustrated in Figure \ref{fig:dataset_generation}.

\section{Experiments}

\subsection{Experimental Settings}
We evaluate models on pre-generated scenarios across the training, validation, and test scenes.
The training scenes consist of two episodes per scene, totaling 1,600 episodes, while the validation and test scenes each consist of 20 episodes per scene, totaling 1,000 episodes.
Each episode comprises $N_\text{goal}$=32 goal positions sampled with geodesic distances in $[d_\text{min}, d_\text{max}]$, where $d_\text{min}=4$ m and $d_\text{max}=8$ m.
We select the robot with a radius of $r_\text{robot}$=0.18 m, a height of $h_\text{robot}$=1.0 m, and a camera height of $h_\text{camera}$=0.6 m, and set the image resolution to $W$ = 256, $H$ = 256.
The action space uses a forward step distance of $d_\text{step}=0.25$ m and a turning angle of $\theta_\text{step}=30^\circ$, following the Habitat Challenge~\cite{habitat2020sim2real}.

\begin{table*}[t]
    \caption{RVN-Bench Results.}
    \label{table:rvnbench_results}
    \centering
    \begin{tabular}{l|l|ccc|ccc|ccc}
    \toprule
    \multirow{2}{*}{Category} & \multirow{2}{*}{Method} & \multicolumn{3}{c|}{Train} & \multicolumn{3}{c|}{Validation} & \multicolumn{3}{c}{Test} \\
        & & SR$_1$ $\uparrow$& $E(G)$ $\uparrow$ & CPK $\downarrow$ & SR$_1$ $\uparrow$& $E(G)$ $\uparrow$ & CPK $\downarrow$ & SR$_1$ $\uparrow$& $E(G)$ $\uparrow$ & CPK $\downarrow$ \\
    \midrule
    \multirow{3}{*}{IL}       & ViNT-PointGoal \cite{shah2023vint}        & 0.097 & 0.11 & 442.3& 0.090 & 0.11 & 468.3& 0.093 & 0.10 & 465.4 \\
                              & NoMaD-PointGoal \cite{sridhar2024nomad}   & 0.758 & 4.01 & 32.5 & 0.752 & 3.69 & 35.7 & 0.751 & 4.52 & 31.0 \\
                              & NoMaD-Neg                                 & 0.750 & 3.83 & 31.2 & 0.757 & 3.97 & 28.6 & 0.760 & 4.61 & 25.8 \\
    \midrule
    \multirow{1}{*}{Safe-RL}  & PPO-Lagrangian \cite{ray2019benchmarking} & 0.826 & 7.88 & 15.7 & 0.803 & 7.61 & 16.9 & 0.805 & 9.02 & 13.7 \\
    \midrule
    \multirow{2}{*}{RL}       & PPO \cite{schulman2017proximal}           & 0.824 & 8.70 & 15.6 & 0.808 & 8.54 & 16.4 & 0.819 & 8.68 & 15.5 \\
                              & DD-PPO \cite{wijmans2020dd}                & 0.899   & 13.28  & 9.2  & 0.865 & 11.86 & 10.5 & 0.886 & 13.90 & 8.7 \\
                              & DDPPO-DAV2 \cite{wijmans2020dd, yang2024dav2}& \textbf{0.937}  & \textbf{20.79}  & \textbf{3.5}  & \textbf{0.909} & \textbf{19.88} & \textbf{4.0} & \textbf{0.928} & \textbf{20.79} & \textbf{3.6} \\
    \bottomrule
    \end{tabular}
\end{table*}

\subsection{Baselines}
In our experiments, we evaluate a diverse set of baseline methods in RVN-Bench. For IL, we include ViNT-PointGoal~\cite{shah2023vint} and NoMaD-PointGoal~\cite{sridhar2024nomad}. For RL, we include PPO~\cite{schulman2017proximal}, DD-PPO~\cite{wijmans2020dd}, and an extended variant, DDPPO-DAV2, which augments RGB input with depth maps estimated by the pre-trained Depth Anything V2 (DAV2) model~\cite{yang2024dav2}. For Safe-RL, we include PPO-Lagrangian~\cite{ray2019benchmarking}. In addition, we include NoMaD-Neg, our proposed baseline that leverages negative trajectory image datasets, which can be collected in RVN-Bench.

\textbf{IL methods.}
ViNT-PointGoal and NoMaD-PointGoal are variants of ViNT \cite{shah2023vint} and NoMaD \cite{sridhar2024nomad} that take the relative goal position as input instead of a goal image.
Because this low-dimensional input has no spatial structure, an image encoder such as EfficientNet is unsuitable and unnecessarily complex for goal encoding. We therefore replace the goal encoder with a lightweight two-layer MLP that maps the goal position to a feature embedding, which is then concatenated with the observation embedding for policy learning. The observation image encoder remains unchanged from the original architectures.
Since ViNT and NoMaD output trajectories rather than discrete actions, the action of the ViNT-PointGoal and NoMaD-PointGoal agents is selected as the discrete action that best follows the lookahead point chosen from the predicted trajectory.
Hyperparameters are set following the official implementations of ViNT and NoMaD.

\begin{figure}[t]
    \centering
    \includegraphics[width=\linewidth]{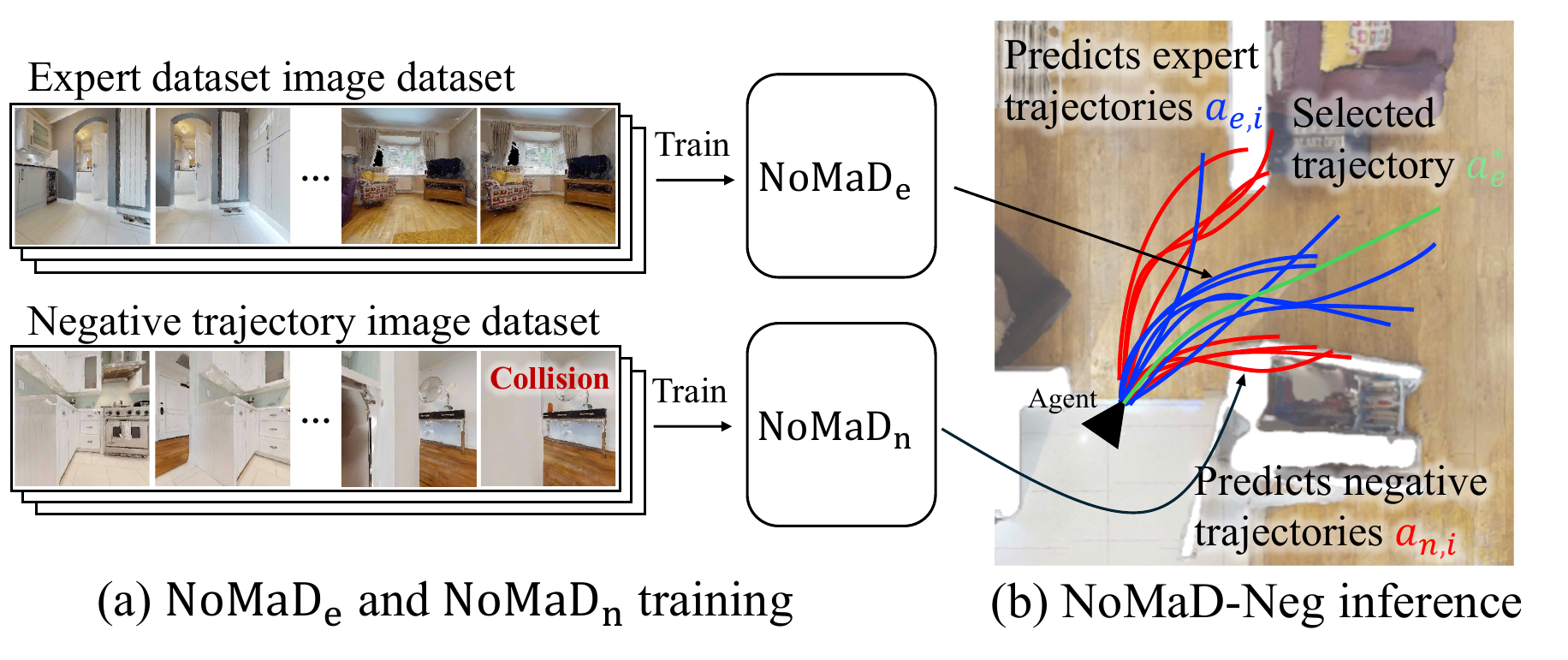}
    \caption{\textbf{Overview of the NoMaD-Neg framework.} (a) Two NoMaD-PointGoal model, NoMaD$_e$ and NoMaD$_n$, are trained separately with expert and negative trajectory datasets. (b) NoMaD$_e$ and NoMaD$_n$ predict expert trajectories $a_{e,i} \in \textbf{a}_E$ and negative trajectories $a_{n,i} \in \textbf{a}_N$, respectively. The expert trajectory $a_e^*$ with minimal $\text{CoR}(a_{e,i}, \textbf{a}_E, \textbf{a}_N)$ is then selected.}
    \label{fig:nomad_neg}
\end{figure}

To take advantage of RVN-Bench’s ability to collect negative trajectory image dataset, we introduce NoMaD‑Neg, a baseline framework extending NoMaD‑PointGoal by leveraging both expert and negative datasets, as illustrated in Fig.~\ref{fig:nomad_neg}.
Specifically, we train two separate NoMaD‑PointGoal models: NoMaD$_e$, trained on the expert dataset to infer collision‑free trajectories toward a given goal as in the original NoMaD‑PointGoal, and NoMaD$_n$, trained on the negative dataset to predict trajectories that collide with obstacles while moving toward the goal. Since NoMaD produces actions probabilistically via a diffusion policy, both NoMaD$_e$ and NoMaD$_n$ can generate diverse trajectories given the same observation. At each time step, NoMaD‑Neg draws $k(=8)$ trajectories $a_{e,i} \in \textbf{a}_E=\{a_{e,0}, ...a_{e,k-1}\}$ from NoMaD$_e$ and $a_{n,i} \in \textbf{a}_N=\{a_{n,0}, ...a_{n,k-1}\}$ from NoMaD$_n$, where $i=\{0, ...,k-1\}$. For each $a_{e,i}$, the constrained reward \cite{lee2020mixgail} ($\text{CoR}(a_{e,i}, \textbf{a}_E, \textbf{a}_N)$) is computed using Equation (\ref{eq:CoR}).

\begin{equation} \label{eq:CoR}
    \begin{aligned}
        \text{CoR}(a, \textbf{a}_E, \textbf{a}_N) &= {(1 + {d_{\textbf{a}_E} \over \alpha})^{-{\alpha+1\over 2}} \over (1 + {d_{\textbf{a}_E} \over \alpha})^{-{\alpha+1\over 2}} + (1 + {d_{\textbf{a}_N} \over \alpha})^{-{\alpha+1\over 2}}},\\
        d_{\textbf{a}_k} &= \sqrt{\frac{1}{|\textbf{a}_k|}\sum_{x \in \textbf{a}_k}{\|a-x\|_2^2}}, \quad k \in \{E, N\}
  \end{aligned}
\end{equation}

The trajectory $a_e^*$ is selected among the $a_{e,i}$ to minimize the CoR loss as defined in Equation(\ref{eq:a_e_star}). Finally, the discrete action $a_t$ that best follows $a_e^*$ is output by NoMaD‑Neg.

\begin{equation} \label{eq:a_e_star}
    a_e^* = \argmin_{a_{e.i}\in\mathbf{a}_E} \text{CoR}(a_{e,i}, \textbf{a}_E, \textbf{a}_N)
\end{equation}

\textbf{RL and Safe-RL methods.}
For PPO \cite{schulman2017proximal}, DD-PPO \cite{wijmans2020dd}, and PPO-Lagrangian \cite{ray2019benchmarking}, we use a ResNet-50 encoder combined with a two-layer LSTM (ResNet-LSTM) model, following the architecture used in the DD-PPO PointNav task\cite{wijmans2020dd}.
We adopt the same hyperparameters as reported in \cite{wijmans2020dd} for PPO and DD-PPO. For PPO-Lagrangian the expected cost limit is set to $10^{-4}$.

In addition, we benchmark the same ResNet-LSTM architecture with estimated depth as an additional input modality. Depth maps are estimated from RGB observations using the pre-trained monocular depth estimation foundation model, DAV2 \cite{yang2024dav2}. The model weights of DAV2 remain frozen during both training and inference, and the predicted depth is concatenated with RGB and provided as input to the ResNet-LSTM model, which is trained using the DD-PPO algorithm. We refer to this model as DDPPO-DAV2.

\textbf{Training.}
For the IL baselines (ViNT-PointGoal, NoMaD-PointGoal), we use the official hyperparameters and train for 100 epochs, increasing the training set size until performance saturates. Training data are collected in RVN-Bench from 800 scenes with 400 trajectories per scene (320k total; $\sim$828\,h assuming 1.0\,m/s). NoMaD-Neg uses the same trajectory budget, replacing 6.25\% of expert trajectories with negative ones. RL and Safe-RL baselines are trained with the hyperparameters of \cite{wijmans2020dd} until convergence. All methods are trained on a single NVIDIA L40S GPU, except DD-PPO and DDPPO-DAV2, which use 8 L40S GPUs.

\subsection{Metrics}
In RVN-Bench, since a new goal is provided within an episode upon reaching the current goal, we use variants of success rate as metrics. 
Specifically, we report the success rate of reaching the first goal ($\text{SR}_{1}$). Additionally, we use the average number of goals reached per episode ($E(G)$) and the number of collisions per kilometer traveled ($\text{CPK}$) as evaluation metrics.

\subsection{RVN-Bench Results}

The experimental results on RVN-Bench are presented in Table \ref{table:rvnbench_results}. 
These results indicate that the collision-aware visual navigation problem remains unsolved. 
Among the baselines, the overall best-performing method was DDPPO-DAV2. While DD-PPO has demonstrated nearly perfect SR and SPL in unseen simulation environments for visual navigation tasks that do not consider collisions \cite{wijmans2020dd}, the algorithm's performance on the RVN-Bench test scenes was notably lower, with an $\text{SR}_{1}$ of 0.928 in the test scene. This highlights the significantly greater difficulty of collision-aware visual navigation and suggests that it remains an open research problem.

Incorporating predicted depth in DDPPO-DAV2 improved over DD-PPO, yielding 1.67× and 1.49× higher $E(G)$ on the validation and test sets, respectively, and reducing CPK by 62\% and 59\%. This indicates that foundation-model depth estimates can significantly improve reactive visual navigation.

Among the baseline methods, all RL-based methods outperformed the IL methods across all metrics. Notably, despite the ResNet-LSTM model used in the RL methods having fewer parameters (12M) than the ViNT-PointGoal (23M) and NoMaD-PointGoal (13M) models, which are used in IL methods, it achieved superior performance. This suggests that learning through interaction with the environment substantially contributes to performance improvements.

NoMaD-Neg, a framework that incorporates a negative dataset into NoMaD-PointGoal, outperformed NoMaD-PointGoal across all metrics in the validation and test scenes. This demonstrates that leveraging both expert and negative datasets can improve sample efficiency and effectively teach the agent to avoid collisions in unseen environments. However, as noted above, NoMaD-Neg still underperformed compared to all RL-based methods across all metrics.

Finally, the evaluation results on the validation and test scenes, which are unseen during training, showed that $\text{SR}_1$ was on average approximately 1.5\% lower than the results on the training scenes. This suggests that models trained on the 800 training scenes of RVN-Bench generalize well to unseen environments.

\begin{table*}[t]
    \caption{Ablation study on effect of observation types of DDPPO.}
    \label{table:ablation_gt_depth}
    \centering
    \begin{tabular}{l|l|ccc|ccc|ccc}
    \toprule
    \multirow{2}{*}{Method} & \multirow{2}{*}{Observation Type}
        & \multicolumn{3}{c|}{Train} 
        & \multicolumn{3}{c|}{Validation} 
        & \multicolumn{3}{c}{Test} \\
        & 
        & SR$_1$ $\uparrow$ & $E(G)$ $\uparrow$ & CPK $\downarrow$ 
        & SR$_1$ $\uparrow$ & $E(G)$ $\uparrow$ & CPK $\downarrow$ 
        & SR$_1$ $\uparrow$ & $E(G)$ $\uparrow$ & CPK $\downarrow$ \\
    \midrule
        \multirow{3}{*}{DD-PPO \cite{wijmans2020dd}}
            & RGB            & 0.899          & 13.28          & 9.2          & 0.865          & 11.86          & 10.5         & 0.886          & 13.90          & 8.7          \\
            & RGB + Predicted Depth \cite{yang2024dav2} 
                             & 0.937          & 22.70          & 3.5          & 0.909          & 19.88          & 4.0          & 0.928          & 20.79          & 3.6          \\
            & RGB + GT Depth & \textbf{0.946} & \textbf{20.79} & \textbf{2.0} & \textbf{0.920} & \textbf{21.90} & \textbf{2.2} & \textbf{0.940} & \textbf{22.99} & \textbf{1.8} \\
    \bottomrule
    \end{tabular}
\end{table*}

\subsection{Real-World Performance of Model Trained on Simulation-Collected Data}

To evaluate whether visual navigation models trained with data collected in simulation can generalize effectively to real-world scenarios, we conducted a real-world experiment in an unseen indoor office and house environment using the NoMaD-PointGoal model. We selected the NoMaD-PointGoal model because it achieved the best performance on RVN-Bench among the methods that can be trained with existing real-world datasets. We trained the NoMaD-PointGoal model with three types of datasets: (1) a real-world dataset comprising Go Stanford \cite{hirose2019gostanford}, RECON \cite{shah2021recon}, SCAND \cite{karnan2022scand} and SACSON \cite{hirose2023sacson} datasets, totaling 54 hours of navigation data; (2) a simulation dataset collected in RVN-Bench as described in Section \ref{traj_data_generation}, totaling 828 hours of navigation data; and (3) the union of the real-world and the simulation-collected datasets. The real-world dataset is a subset of the ViNT training dataset \cite{shah2023vint} and includes both indoor office datasets \cite{hirose2019gostanford, hirose2023sacson} and dataset collected using the Jackal UGV \cite{clearpath_jackal, shah2021recon, karnan2022scand}.

\begin{figure}[t]
    \centering
    \includegraphics[width=\linewidth]{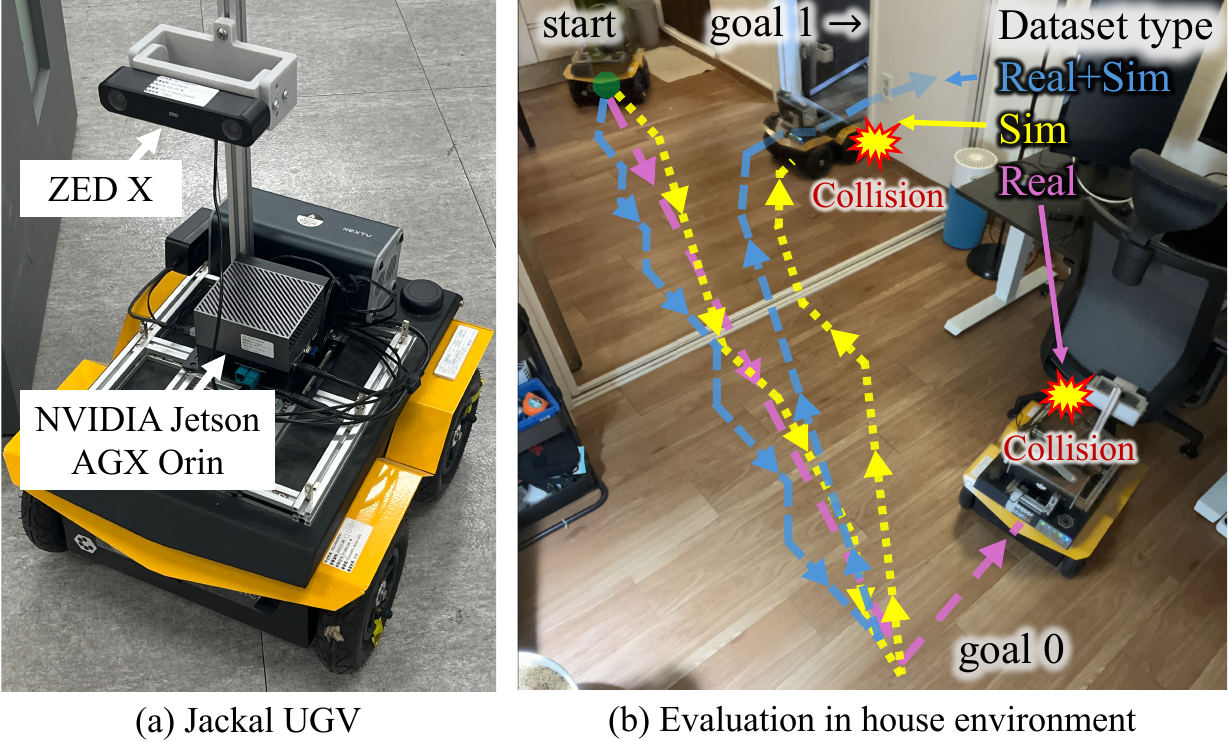}
    \caption{\textbf{Real-world evaluation of NoMaD-PointGoal with different training datasets.} (a) Clearpath Jackal UGV platform  \cite{clearpath_jackal, stereolabs_zedx, nvidia_jetson_orin} used in the experiments. (b) Illustration of evaluation results in a house environment. Shown are the executed trajectories of the NoMaD-PointGoal agent trained on the real-world dataset (pink), trained on the simulation dataset (yellow), and trained on the combined real-world and simulation datasets (blue).}
    \label{fig:real_exp}
\end{figure}

The real-world experiments were conducted using a Jackal UGV \cite{clearpath_jackal} platform shown in Figure \ref{fig:real_exp} (a).
The Jackal UGV platform was equipped with a ZED-X 2.2mm camera \cite{stereolabs_zedx} mounted at a height of $h_\text{camera}$=0.6 m, consistent with the RVN-Bench configuration.
All inference was performed onboard using an NVIDIA Jetson AGX Orin computer \cite{nvidia_jetson_orin}.
The action space $\mathcal{A}$ of the real robot followed the same discrete action set defined in the RVN-Bench in Section \ref{taskdef}.
The goal was considered successfully reached if the robot approached within 0.36 m of the goal position.
A total of 20 evaluation episodes were conducted, each with $N_g=5$, comprising 15 episodes in office environments and 5 in house environments.

\begin{table}[t]
    \caption{Real-world evaluation results of NoMaD-PointGoal models trained with different training datasets}
    \label{table:nomad_real}
    \centering
    \begin{tabular}{l|l|ccc}
    \toprule
    Model & Dataset & SR$_1$ $\uparrow$& $E(G)$ $\uparrow$ & CPK $\downarrow$ \\
    \midrule
    \multirow{3}{*}{NoMaD-PointGoal\cite{sridhar2024nomad}} & Real     & 0.30 & 0.30 & 223.5 \\
                                                            & Sim      & 0.60 & 1.05 & 196.4 \\
                                                            & Real+Sim & \textbf{0.75} & \textbf{1.30} & \textbf{191.4}  \\
    \bottomrule
    \end{tabular}
\end{table}

The results of this evaluation are summarized in Table \ref{table:nomad_real} and example executed trajectories in a house environment are shown in Figure \ref{fig:real_exp} (b).
In the illustrated episode, all three agents successfully reached the first goal position.
However, the agent trained on the real-world dataset collided with a chair, and the agent trained on the simulation dataset collided with a wall. In contrast, the agent trained on the combined real-world and simulation datasets navigated to the next room, where the subsequent goal position was located, without collision.

The model trained solely on simulation-collected data demonstrated strong generalization capabilities, achieving higher success rates and fewer collisions compared to the models trained exclusively on real-world dataset. In particular, the simulation-trained model achieved a 3.5× improvement in $E(G)$ over the real-only model, while the model trained with both simulation and real-world data achieved a 4.3× improvement in $E(G)$. Both the simulation-trained and hybrid models also exhibited substantially lower CPK than the real-only model. Incorporating both real-world and simulation dataset yielded the best overall performance, suggesting that large-scale simulation datasets collected in RVN-Bench can complement limited real-world data and enhance generalization to real-world deployment.

\subsection{Ablation Study: Effect of Depth in Reactive Navigation}

To evaluate the impact of incorporating ground-truth depth information in the reactive navigation task, we compare the performance of the DD-PPO model across different observation modalities. The results of DD-PPO trained with RGB, RGB plus predicted depth, and RGB plus ground-truth depth are presented in Table \ref{table:ablation_gt_depth}. Incorporating ground-truth depth alongside RGB achieves the strongest results, yielding an average 10.2\% improvement in $E(G)$ over the model using predicted depth on the validation set and 10.6\% improvement on the test set.

Furthermore, CPK decreased by approximately 45\% when the model uses ground-truth depth compared to predicted depth on the validation set and 50\% on the test set.
These findings suggest that continued advances in monocular depth estimation are likely to yield further performance gains in reactive visual navigation.

\section{Conclusion and Future Work}

We presented RVN-Bench, a collision-aware benchmark for reactive visual navigation in indoor environments. Unlike existing benchmarks that focus solely on goal reaching and ignore collisions, RVN-Bench explicitly evaluates both success and safety.
It provides standardized environments for reinforcement learning and a pipeline for generating trajectory image datasets, including negative trajectories that are costly to collect in the real world.
Our experimental results demonstrated that RVN-Bench poses a significantly greater challenge than existing visual navigation benchmarks.
Preliminary real-world experiments showed promising sim-to-real transfer, though the evaluation was limited in scale.
RVN-Bench is limited to static environments, discrete action spaces, and NavMesh-based kinematics, and our physical evaluation was conducted on a single hardware platform. Future work will expand this scope by incorporating dynamic obstacles, continuous control, and broader robot support.






\bibliographystyle{IEEEtran}
\bibliography{main}

\end{document}